\documentclass{ifacconf}
\usepackage{xcolor}
\usepackage{svg}
\usepackage{graphicx} % 
\usepackage{svg} % 
\usepackage{subcaption} %
\usepackage{booktabs}
\usepackage{amsmath}

\newtheorem{lemma}{Lemma}[section]

\usepackage{amsfonts}

\usepackage{graphicx}      % include this line if your document contains figures
\usepackage{natbib}        % required for bibliography
\begin{document}
\begin{frontmatter}

\title{A Navigation System for ROV's inspection on Fish Net Cage\thanksref{footnoteinfo}} 

\thanks[footnoteinfo]{This research was financially supported by start funding of 100 talent program at ZJU-Hangzhou Global Scientific and Technological Innovation Center}
\author[First]{$^\dagger$Zhikang Ge} 
\author[Second]{$^\dagger$Fang Yang} 
\author[First]{Wenwu Lu} 
\author[Third]{Peng Wei}
\author[Second]{Yibin Ying} 
\author[First]{Chen Peng}
\thanks{Corresponding author: Chen Peng, e-mail: chen.peng@zju.edu.cn}
\thanks{$\dagger$ authors are equally contributed to this paper}

\address[First]{ZJU-Hangzhou Global Scientific and Technological Innovation Center, Zhejiang University, Hangzhou, 311215, China }
\address[Second]{College of Biosystems Engineering and Food Science, Zhejiang University, Hangzhou, 310058, China }
\address[Third]{Department of Biological and Agricultural Engineering, University of California, Davis, 95618, USA }

\begin{abstract}% Abstract of not more than 250 words.
Autonomous Remotely Operated Vehicles (ROVs) offer a promising solution for automating fishnet inspection, reducing labor dependency, and improving operational efficiency. In this paper, we modify an off-the-shelf ROV, the BlueROV2, into a ROS-based framework and develop a localization module, a path planning system, and a control framework. For real-time, local localization, we employ the open-source TagSLAM library. Additionally, we propose a control strategy based on a Nominal Feedback Controller (NFC) to achieve precise trajectory tracking. The proposed system has been implemented and validated through experiments in a controlled laboratory environment, demonstrating its effectiveness for real-world applications.
\end{abstract}


\end{frontmatter}
%===============================================================================

\section{Introduction}
Aquaculture is a critical component of global food production, with significant potential to enhance food security, nutrition, and contribute to economic growth and environmental sustainability (\cite{FAO:24}). However, as the industry rapidly expands, it faces substantial challenges in adopting sustainable and environmentally responsible practices. One major concern is fish escape, often caused by net damage (\cite{fore:21}). Additionally, biofouling on net enclosures can reduce oxygen levels, negatively impacting fish health (\cite{ohrem:20}). Biofouling and net damage are common issues in sea-based fish farms, making regular net inspections essential for identifying these problems and guiding maintenance and cleaning efforts to ensure safe and sustainable fish production.

Traditional manual inspection methods rely on divers, which involve high safety risks and inefficiency. Aquaculture robotics are addressing net pen inspection challenges, improving safety and efficiency (\cite{vasileiou:24}). These robots are capable of efficiently performing inspection tasks in complex underwater environments, reducing human intervention and enhancing safety. One of the key challenges for underwater robots is high-precision positioning and navigation (\cite{AmuXanFor:24}). Since traditional GPS cannot be used underwater, robots rely on sonar, depth sensors, and computer vision technologies for real-time localization and environmental perception. With these technologies, robots can autonomously detect net damage, deformation, and biofouling, ensuring thorough and accurate inspections.

Underwater inspection of fishnets in aquaculture has seen significant advancements with the development of various autonomous underwater vehicles (AUVs) and remotely operated vehicles (ROVs). For instance, an AUV system designed by \cite{stenius:22} for seaweed farm inspections employs dead-reckoning and sonar for localization, but it does not account for dynamic environmental disturbances. In a similar approach, \cite{akram:22} proposed a vision-based servoing system for ROVs that combines object detection and closed-loop control to track net pens. However, this method proves sensitive to underwater visibility, which can limit its effectiveness. Meanwhile, \cite{betancourt:20} integrated computer vision with ROVs for inspecting net cages, achieving high accuracy in detecting net patterns, though their system faces challenges from water currents. To improve trajectory tracking during net pen inspections, \cite{cardaillac:23} introduced a maneuvering-based control system for ROVs. This solution, however, assumes relatively stable environmental conditions, which may not always be the case. In GPS-denied navigation, \cite{lopez-barajas:23} developed a system using optical cameras and convolutional neural networks (CNN) to control the distance between the AUV and the net, although this approach is limited to controlled conditions. Similarly, \cite{amundsen:21} created a DVL-based navigation system for ROVs that minimizes error in the presence of currents, but it still requires further field validation. In a different direction, \cite{lopez-barajas:24} demonstrated the use of deep learning for hole detection in nets, showing robust inspection capabilities despite challenges with visibility. \cite{vasileiou:24} designed a cost-effective, 3D-printed AUV for frequent inspections, but its performance in dynamic environments needs further exploration. Finally, \cite{tani:24} proposed a visual-acoustic system for relative navigation, which is effective in dynamic tasks but has yet to be tested in the specific context of aquaculture net pen inspections.

These studies advance autonomous fishnet inspection but still face challenges like environmental disturbances and poor visibility, necessitating multi-sensor integration. Common issues include localization problems due to limited visual features and lighting, and tracking controllers tested without accounting for underwater currents. In aquaculture, fishnet inspections occur in confined spaces near net pens, where these disturbances are more pronounced. As a result, many existing AUV systems and simplified tracking methods are inadequate for these specific tasks. To address this, this paper presents a standard task scenario for fishnet inspection, using a 4-DoF BlueROV2 with six thrusters for trajectory tracking control. Simultaneous Localization and Mapping (SLAM) with Fiducial Markers (\cite{wang2020acmarker}) is used for accurate localization in the featureless environments, and an NFC controller is developed for experimental simulation based on this scenario. In conclusion, the key contributions of the work presented in this paper can be summarized as follows:
 \begin{itemize}
\item We developed an autonomous ROV system for fishnet cage inspection by integrating a commercial BlueROV2 platform within a ROS framework, delivering advanced localization, path planning, and control capabilities.
\item We designed a high-precision localization module using the  Fiducial Markers, achieving real-time, accurate positioning even in featureless underwater environments.
\item We proposed a control strategy based on a Nominal Feedback Controller (NFC) that ensures precise trajectory tracking and robust performance despite uncertainties and disturbances during inspections.
\end{itemize}

The remaining part of this paper is structured as follows. Section 2 provides an overview of the ROV Inspection System, including the robotic platform, localization system, path planning system, and the control system, with a focus on dynamics modeling and the NFC controller. Section 3 presents the experimental setup, followed by the results and analysis in Section 4. Finally, Section 5 concludes the paper and outlines directions for future work.

% \end{itemize}
\section{ROV Inspection System}
In the ROV autonomous inspection system, the effective operation hinges on the seamless integration of three fundamental components: the localization, path planning, and control systems. The localization system continuously tracks the ROV’s position, providing real-time, precise location data, which enables the ROV to navigate and orient itself within the complex underwater environment. The path planning system, leveraging this positional information and task-specific objectives, generates an optimal inspection route. Finally, the control system adjusts the ROV’s movement, continuously monitoring both the execution of the planned trajectory and feedback from the localization system, ensuring stable and accurate performance throughout the inspection process.
\subsection{Robotic Platform}
\begin{figure}[h!]
    \centering
\includegraphics[width=0.85\columnwidth]{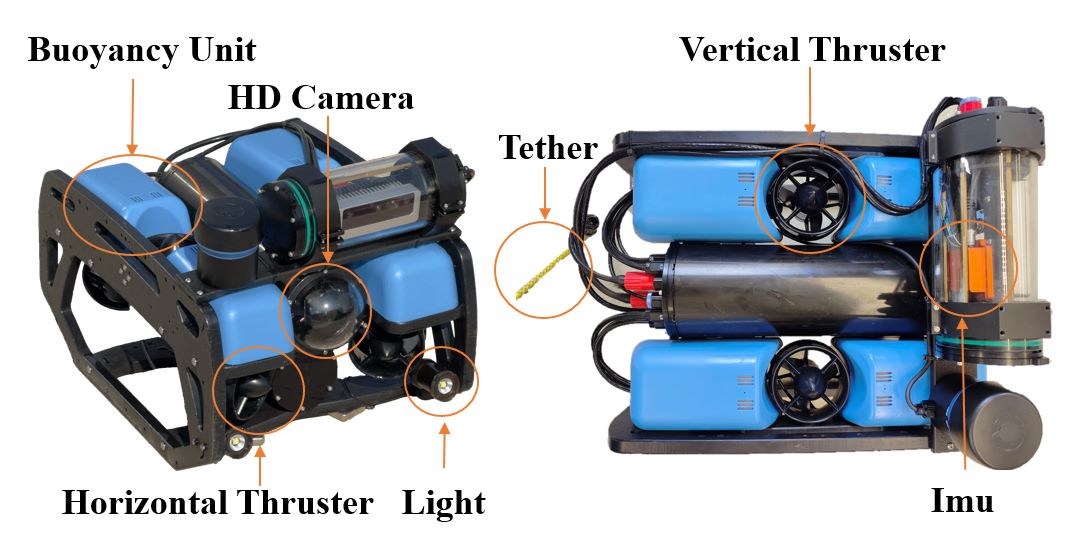}
    \caption{The integrated sensors of the BlueROV2 for the fish net cage inspection.}
    \label{fig:ROV_components}
\end{figure}

Our robotic system is built on the BlueROV2 \cite{bluerov2}, modified for autonomous fishnet inspection by enhancing both hardware and software. Figure~\ref{fig:ROV_components} displays the BlueROV2 and its integrated sensors. The integrated sensors and components are depicted in Figure~ref{fig:Software and hardwareframeworkof the ROV}.
On the hardware side, we added additional sensors, such as the IMU, stereo camera, and the original HD camera, to improve localization and support precise inspection tasks. These sensors provide critical data for navigation and net damage detection, ensuring stable operation.

For the low-level ROV system, we reinstalled the original setup and deployed the Ubuntu-ROS operating system, integrating it with the control system to upload sensor data and execute motion commands from the host computer. This modification enables autonomous operation while maintaining a reliable communication link for real-time control and monitoring of the fishnet inspection process.

\begin{figure}[h!]
    \centering
    \includegraphics[width=1\columnwidth]{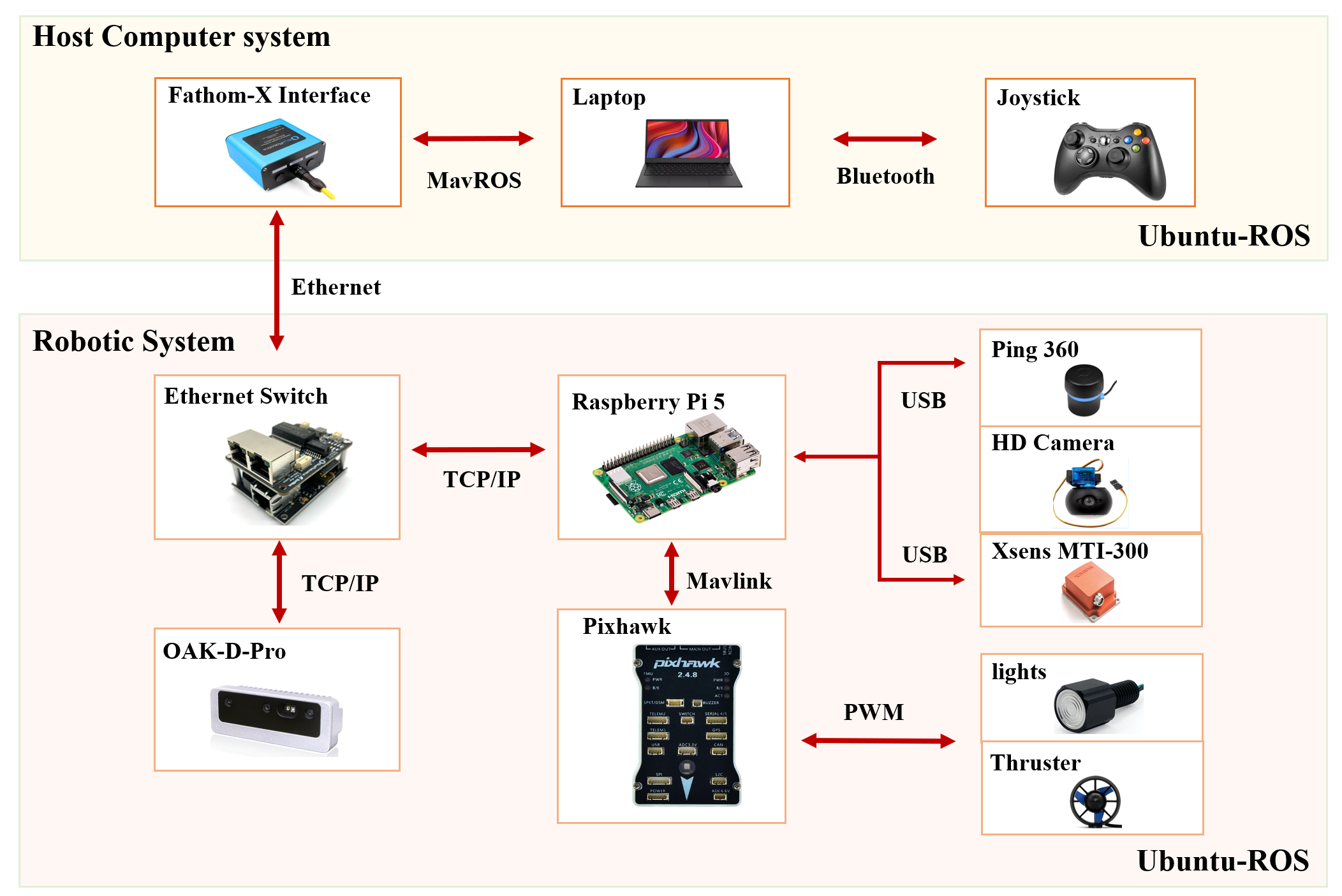}
    \caption{ROV with the integrated sensor setup.}
    \label{fig:Software and hardwareframeworkof the ROV}
\end{figure}

\subsection{Localization system}
\label{sec::robot_localization}
To address the challenges of limited global localization and sparse visual features in underwater environments, we employed the open-source TagSLAM library by ~\cite{Pfrommer:19} for real-time localization. The library integrates marker detection, pose estimation, and optimization to deliver reliable localization, even in dynamic environments with multiple markers.

\begin{figure}[h]
\centering
\includegraphics[width=0.45\linewidth]{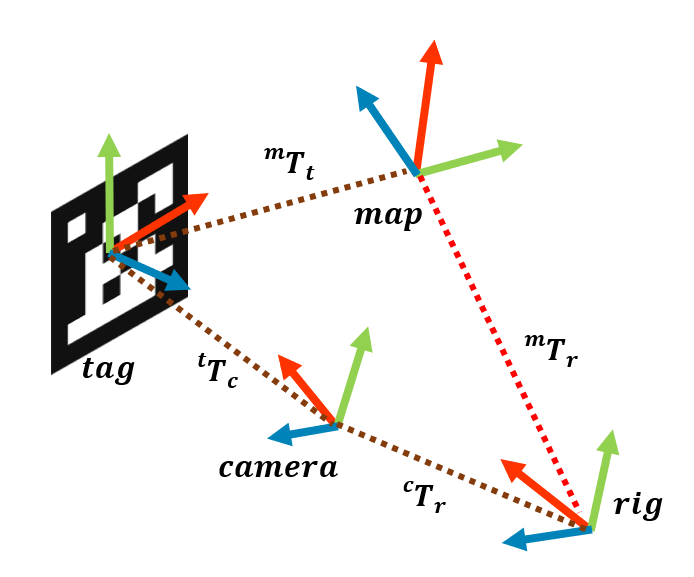}
\caption{A TagSLAM scene with a single camera, featuring dynamic "rig" and a static "map" body.}
\label{fig::Single-camera TagSLAM scene}
\end{figure}

As illustrated in Figure~\ref{fig::Single-camera TagSLAM scene}, TagSLAM enables the modeling of various SLAM problems by abstracting the roles of bodies, tags, and cameras. The key to calculating the robot’s position lies in understanding the transformations between coordinate systems. Specifically, the goal is to compute the pose of the robot body frame (rig) relative to a pre-built map. 
% Using the following equations, we can achieve this:

% \begin{equation} 
% \mathbf{B_X} = T_A^B \cdot \mathbf{A_X} 
% \end{equation}

% In this equation, \( \mathbf{A_X} \) is the position vector of a point in coordinate frame \( A \), and \( T_A^B \) is the transformation matrix that converts coordinates from frame \( A \) to frame \( B \), giving us the point's position \( \mathbf{B_X} \) in frame \( B \).

% \begin{equation} 
% T_{\text{map}}^{\text{rig}} = T_{\text{map}}^{\text{tag}} \cdot T_{\text{tag}}^{\text{camera}} \cdot T_{\text{camera}}^{\text{rig}} 
% \end{equation}

% In this second equation, the goal is to compute the transformation \( T_{\text{map}}^{\text{rig}} \), which gives the rig's position in the map frame. To do so, we combine the following known transformations: 
Based upon the observation of the camera on the tag, we can have an online estimation of the camera pose relative to the observed marker. Given the pose of the tag in the pre-built map, we can have an estimation of the robot body in the map, which can be expressed as the chain rule of the coordinate transformation (see Figure~\ref{fig::Single-camera TagSLAM scene}).

\begin{equation} 
T_{\text{map}}^{\text{rig}} = T_{\text{map}}^{\text{tag}} \cdot T_{\text{tag}}^{\text{camera}} \cdot T_{\text{camera}}^{\text{rig}} 
\end{equation}

% \begin{itemize}
%     \item \( T_{\text{map}}^{\text{tag}} \), the transformation that describes the tag's position in the map frame
%     \item \( T_{\text{tag}}^{\text{camera}} \), the transformation between the tag and the camera frame
%     \item \( T_{\text{camera}}^{\text{rig}} \), the transformation between the camera and the rig frame
% \end{itemize}

% By combining these transformations, we can accurately compute the rig’s pose within the map, ensuring precise localization. This method is robust, supporting loop closure in odometry and handling environments with sparse visual features.

\subsubsection{Mapping with Markers}
The localization map was generated by manually piloting the robot throughout the environment to allow TagSLAM to generate an initial spatial map. This mapping operation is carried out only once, with the resulting map saved locally for use in all subsequent experiments, thereby eliminating the need for repeated mapping. 

\subsubsection{Sensor Fusion for Accuracy}
To enhance localization accuracy, the odometry information from TagSLAM was integrated  with inertial data from the onboard IMU via an Extended Kalman Filter (\cite{MooreStouchKeneralizedEkf2014}). In this way, the robot pose in four controlled degrees are estimated online with higher frequency and precision.

\subsection{Path Planning system}
\begin{figure}[h!]
    \centering
    \includegraphics[width=1\columnwidth]{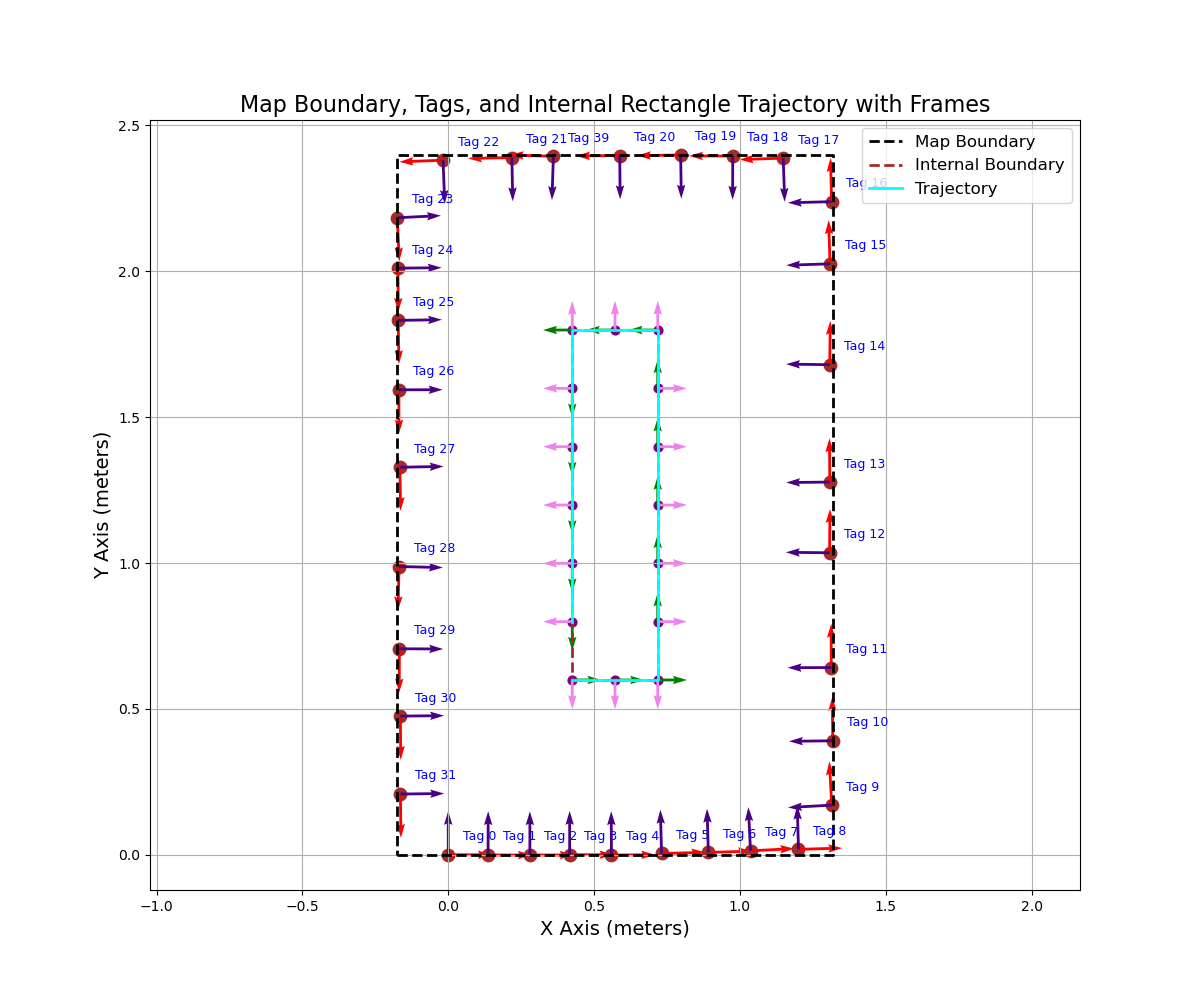} % 
    \caption{The planner of ROV's inspection given a constructed map.} 
    \label{fig:maptagtra} 
\end{figure}

\begin{figure}[h!]
    \centering
    \includegraphics[width=0.85\columnwidth]{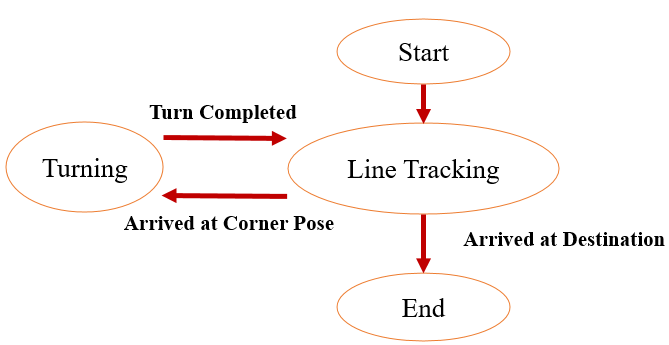} % 
    \caption{Motion States and Reference Frames of the ROV.} 
    \label{fig:Planning_flowchart} 
\end{figure}

During the path planning phase, the global map of the net boundary, previously constructed, is represented as a closed polygon. From this, an offset path is generated by shrinking the boundary curve inward by a specified margin, ensuring an appropriate stand-off distance for the inspection. This strategy guarantees that the ROV follows a net-facing trajectory while staying controllable distance to conduct effective visual or sensor-based assessments of the net, as illustrated in Figure~\ref{fig:maptagtra}.
To handle the transitions between various navigational and inspection tasks, a finite state machine (FSM) is implemented, as shown in Figure~\ref{fig:Planning_flowchart}. Each state within the FSM represents a specific operational mode, from basic waypoint-following to corner turning. The FSM framework enables the system to dynamically switch control strategies based on different navigation states, ensuring reliable and adaptable performance in the dynamic aquaculture environment.

\subsection{Control System}
\label{sec::robust_controller_design}

\subsubsection{Dynamics Modeling}

\begin{figure}[h!]
    \centering
    \includegraphics[width=0.5\columnwidth]{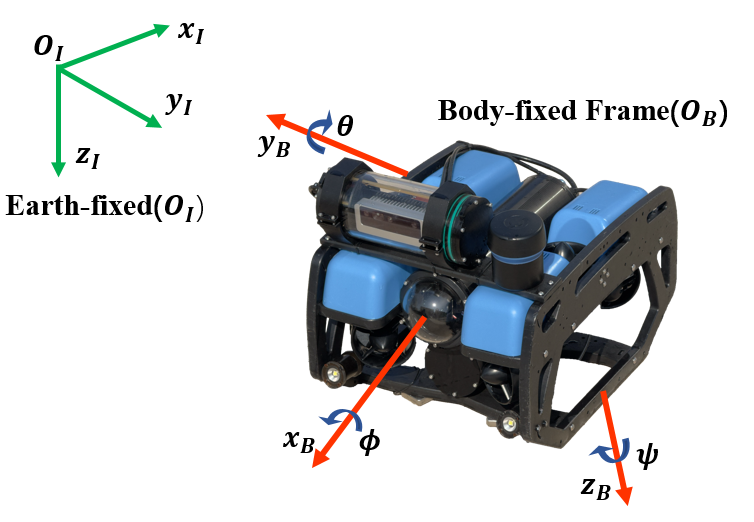} % 
    \caption{The reference frames of an UUV include the Earth-fixed frame(\(O_I\)) and the body-fixed frame (\(O_B\)).} 
    \label{fig:ROV_model} 
\end{figure}

Fig.~\ref{fig:ROV_model} defines the body-fixed frame {$\mathcal{O_B}$} of the applied BlueROV2 and the North-East-Down (NED) reference frame {$\mathcal{O_I}$} used to describe its kinimetics. To simplify the model, the following assumptions are made following this paper by \cite{haugalokken2023adaptive}.

The ROV dynamic model with 4 Degrees-of-Freedom(DoF) is given by \cite{fossen:11}:

\begin{equation}
M \dot{\nu} + C(\nu) \nu + D(\nu) \nu + g +\tau_d = \tau 
\label{eq:rov_dynamic_1_body}
\end{equation}

\begin{equation}
\dot{\eta} = J(\psi) \nu
\label{eq:rov_dynamic_2_body}
\end{equation}

where $\nu = [u, v, w, r]^T$ is the velocity vector in $\mathcal{O_B}$, and $\eta = [x, y, z, \psi]^T$ represents the vector of position and orientation in $\mathcal{O_I}$. $M \in \mathbb{R}^{4 \times 4}$ is the total inertial matrix accounting for both the rigid-body inertia and the hydrodynamic added mass effects, defined as in Eq.\ref{eq:3}. $C(\nu) \in \mathbb{R}^{4 \times 4}$ is the Coriolis matrix, which contains the centripetal and Coriolis forces arising from both rigid-body and added mass contributions, denoted as Eq.\ref{eq:4}. $D(\nu) \in \mathbb{R}^{4 \times 4}$ is the hydrodynamic damping matrix composed of both linear and nonlinear components as given by Eq.\ref{eq:5}.
\begin{equation}
M = \begin{bmatrix}
m_{11} & 0 & 0 & 0 \\
0 & m_{22} & 0 & 0 \\
0 & 0 & m_{33} & 0 \\
0 & 0 & 0 & m_{44}
\end{bmatrix}
\label{eq:3}
\end{equation}

\begin{equation}
C(\nu) = \begin{bmatrix}
0 & 0 & 0 & -m_{22} v \\
0 & 0 & 0 & m_{11} u \\
0 & 0 & 0 & 0 \\
m_{22} v & -m_{11} u & 0 & 0
\end{bmatrix}
\label{eq:4}
\end{equation}

\begin{equation}
D(\nu) = \left[
\begin{array}{@{\hskip 0in}c@{\hskip -0.1in}c@{\hskip -0.1in}c@{\hskip -0.1in}c@{\hskip -0.01 in}c}
d_{l1} + d_{n1} |u| & 0 & 0 & 0 \\
0 & d_{l2} + d_{n2} |v| & 0 & 0 \\
0 & 0 & d_{l3} + d_{n3} |w| & 0 \\
0 & 0 & 0 & d_{l4} + d_{n4} |r|
\end{array}
\right]
\label{eq:5}
\end{equation}

$g= [0, 0, \mathcal{B}-\mathcal{W}, 0]^T$ is the vector of hydrostatic forces, where $\mathcal{B}-\mathcal{W}$ represents the resultant force of self-gravity in the water. $\tau_d = \begin{bmatrix} \tau_u, \tau_v, \tau_w, \tau_r \end{bmatrix}^T$ represents the vector of forces and moments in $\mathcal{O_B}$. The disturbance vector is denoted as $\tau = \begin{bmatrix} \tau_{d1}, \tau_{d2}, \tau_{d3}, \tau_{d4} \end{bmatrix}^T$. Finally, the transformation matrix, $J(\psi) \in \mathbb{R}^{4 \times 4}$, from the body-fixed frame $\mathcal{O_B}$ to the world frame $\mathcal{O_I}$ is given by:
\begin{equation}
J(\psi) = 
\begin{bmatrix}
\cos(\psi) & -\sin(\psi) & 0 & 0 \\
\sin(\psi) & \cos(\psi) & 0 & 0 \\
0 & 0 & 1 & 0 \\
0 & 0 & 0 & 1 
\end{bmatrix}
\label{eq:6}
\end{equation}

%% bound of J(\psi)\ 

%It is easily to prove that the norm of the Jacobian matrix \( J(\psi)\) is bounded. Thus, we have a known positive constant such that:
%%%%This ensures that the effects of the transformation matrix remain within a controllable range, regardless of the variations in parameter 

%This notation indicates that the norm (or maximum magnitude) of the Jacobian matrix J(ψ)J(\psi) is limited by an upper bound JJ, regardless of the parameter η\eta.

We can express the dynamic model in Eq~\ref{eq:rov_dynamic_1_body} and Eq~\ref{eq:rov_dynamic_2_body} fully in the world frame $\mathcal{O_I}$ in Eq~\ref{eq:7}.
\begin{equation}
M_\eta \ddot{\eta} + C(\nu, \eta) \eta + D(\nu, \eta) \eta + g_\eta(\eta) + \tau_e = \tau_\eta 
\label{eq:7}
\end{equation}
where
\begin{equation}
\left\{\begin{aligned}
    M_\eta &= J^{-T}(\psi) M J^{-1}(\psi) \\
    C(\nu, \eta) &= J^{-T}(\psi) \left[ C(\nu) - MJ^{-1}(\psi) \dot{J}(\psi) \right] J^{-1}(\psi) \\
    D(\nu, \eta) &= J^{-T}(\psi) D(\nu) J^{-1}(\psi) \\
    g_\eta(\eta) &= J^{-T}(\psi) g \\
    \tau_e &= J^{-T}(\psi) \tau_d  
\end{aligned}\right.
\label{eq:8}
\end{equation}

\subsubsection{NFC Controller}
In this section, NFC is designed to stabilize the ROV's controlling system.

The tracking error of the ROV is defined to quantify the deviation of the actual trajectory from the desired one. Let $\eta_d = [x_d, y_d, z_d, \psi_d]^T$ represent the desired position and yaw angle from the desired trajectory, the tracking error can be denoted as in Eq.~\ref{eq:10}.
\begin{equation}
\varepsilon = \eta - \eta_d = 
\begin{bmatrix}
x - x_d \\
y - y_d \\
z - z_d \\
\psi - \psi_d
\end{bmatrix}
\label{eq:10}
\end{equation}

As the force and torque are input to the low level system of ROV, the second order of the error dynamics are derived by differentiating $\varepsilon$ with respect to time:
\begin{align}
\ddot{\varepsilon} &= \ddot{\eta} - \ddot{\eta}_d \notag \\
&= - M_{\eta}^{-1}\left[C(\nu,\eta) + D(\nu,\eta)\right]\dot{\varepsilon} 
   + M_{\eta}^{-1}\left(\tau_\eta - g(\eta) - \tau_e(\eta) \right) \notag \\
&\quad - M_{\eta}^{-1}\left[C(\nu,\eta) + D(\nu,\eta)\right]\dot{\eta}_d 
   - \ddot{\eta}_d 
\end{align}

We obtain the following the error dynamics of the ROV control system:
\begin{equation}
\dot{e} = A e + B f
\label{eq:error_dynamics}
\end{equation}
where
\begin{equation}
e = \begin{bmatrix} \varepsilon^T & \dot{\varepsilon}^T \end{bmatrix}^T
\end{equation}
\begin{equation}
A = 
\begin{bmatrix}
0 & I \\
0 & N
\end{bmatrix}, \quad
N = -M_{\eta}^{-1}\left[C(\nu,\eta) + D(\nu,\eta)\right]
\end{equation}

\begin{equation}
B = \begin{bmatrix} 0 & I \end{bmatrix}^T 
\end{equation}

\begin{equation}
f = M_{\eta}^{-1}u_1 + f_1
\end{equation}
with
\begin{equation}
f_1 = M_{\eta}^{-1} \left[ -g(\eta) - \tau_e(\eta) - \left(C(\nu,\eta) + D(\nu,\eta)\right)\dot{\eta}_d \right] - \ddot{\eta}_d
\end{equation}

To stabilize the error dynamics in \eqref{eq:error_dynamics}, we present the following lemma:

\begin{lemma}
The error dynamics equation \eqref{eq:error_dynamics} for the control system asymptotically converges to zero if the following nominal feedback control law is used:
\begin{equation}
\tau_\eta = M_{\eta}\left(K e + \ddot{\eta}_d\right) + \left[C(\nu,\eta) + D(\nu,\eta)\right]\dot{\eta}_d  + g(\eta) + \tau_e(\eta)
\label{eq:control_law}
\end{equation}
where $K = [-K_1, -K_2]$, $K_1 \in \mathbb{R}^{4 \times 4}$, $K_2 \in \mathbb{R}^{4 \times 4}$, and matrix $K$ is designed such that
\begin{equation}
A_1 = A + B K
\end{equation}
is an asymptotically stable matrix.
\end{lemma}

% \begin{proof}
The proof was described in detail by \cite{Leitmann:81} and \cite{Singh:86}.
% \end{proof}

\section{Experiments and Results} 
\label{sec::experiments_and_results}
\begin{figure}[h]
\centering
\includegraphics[width=0.8\linewidth]{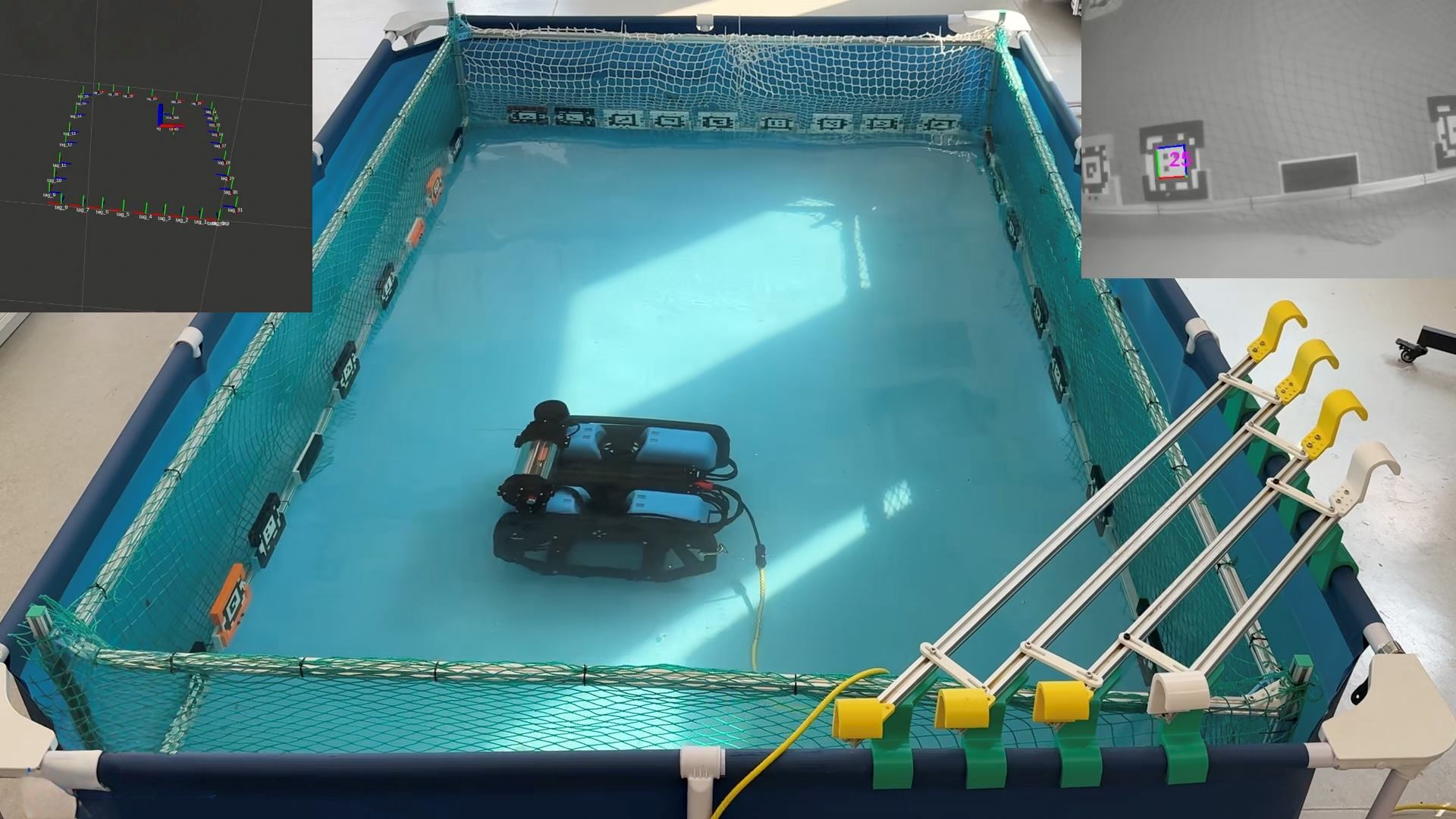}
\caption{The setup of the experiment in the lab ZJU-Hangzhou Global Scientific and Technological Innovation Center in Zhejiang.}
\label{fig::labsetup}
\end{figure}

This section presents laboratory results (as shown in Figure~\ref{fig::labsetup}) evaluating the proposed ROV inspection system for fishnet inspection. The experiments demonstrated the system's stability and precision in a controlled environment, confirming its capability to effectively handle real-world uncertainties and follow complex inspection paths. The horizontal speed (along y-axis of $\mathcal{O_B}$) of ROV is set as 0.1 m/s for the mission of fishnet inspection.

\subsection{Setup and configuration}
We evaluated the performance of the proposed NFC-based control scheme against a classical PID controller in a laboratory water tank (dimensions \(259\,\mathrm{cm} \times 170\,\mathrm{cm} \times 61\,\mathrm{cm}\)). An enhanced BlueROV2 platform, capable of precise four-degree-of-freedom maneuvering, was used alongside a TagSLAM localization system refined by an Extended Kalman Filter for accurate state estimation. 

\subsubsection{PID controller}For the PID controller, which maintained a constant inspection velocity of \(v_y = 0.10\,\text{m/s}\) along the \(y\)-axis, the gains were set as follows: lateral control (\(K_p=400\), \(K_i=0.0\), \(K_d=50\)); yaw control (\(K_p=150\), \(K_i=0.0\), \(K_d=15\)); depth control (\(K_p=500\), \(K_i=0.0\), \(K_d=50\)); and throttle control (\(K_p=350\), \(K_i=0.0\), \(K_d=15\)). 

\subsubsection{NFC controller}The NFC controller was implemented using dynamic model parameters as follows: the mass matrix \( M_0 = \text{diag}([25.2, 25.2, 25.2, 0.402]) \). The Coriolis matrix parameters are set as \( m_{11} = 12.5 \) and \( m_{22} = 12.5 \), and the corresponding Coriolis matrix \( C_0 \) follows the standard 4-DoF formulation. The damping parameters are set as \( Xu = -5.5 \), \( Yv = -7.0 \), \( Zw = -8.0 \), and \( Nr = -1.0 \), and the corresponding damping matrix \( D_0 = -\text{diag}([Xu, Yv, Zw, Nr]) \).  The hydrostatic force vector is assumed to be \( g_0 = [0, 0, 0, 0]^T \) under near-neutral buoyancy conditions. The external disturbance is chosen \( \tau_{e0} = [0, 0, 0, 0]^T \). The control gain matrices were set as \( K_1 = \text{diag}([-300.0, -350.0, -1500.0, -250.0]) \) and \( K_2 = \text{diag}([-100.0, -70.0, -300.0, -60.0]) \).

\subsection{Results and Analysis}
\begin{figure}[h]
\centering
\includegraphics[width=0.8\linewidth]{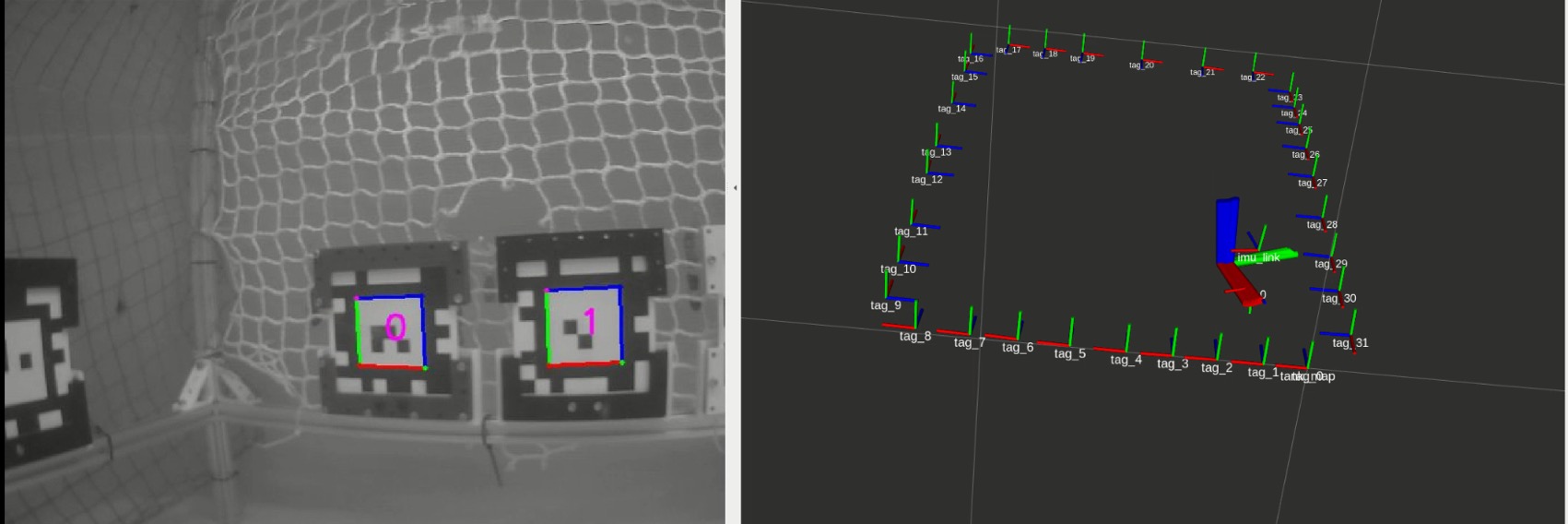}
\caption{Localization results using TagSLAM: left is the image seen by ROV's camera module and right is the localization result in the pre-built map.}
\label{fig::tagslam_localization_result}
\end{figure}

\begin{figure}[h!]
    \centering
    \begin{subfigure}[b]{0.45\textwidth}
        \centering
        \includegraphics[width=0.7\textwidth]{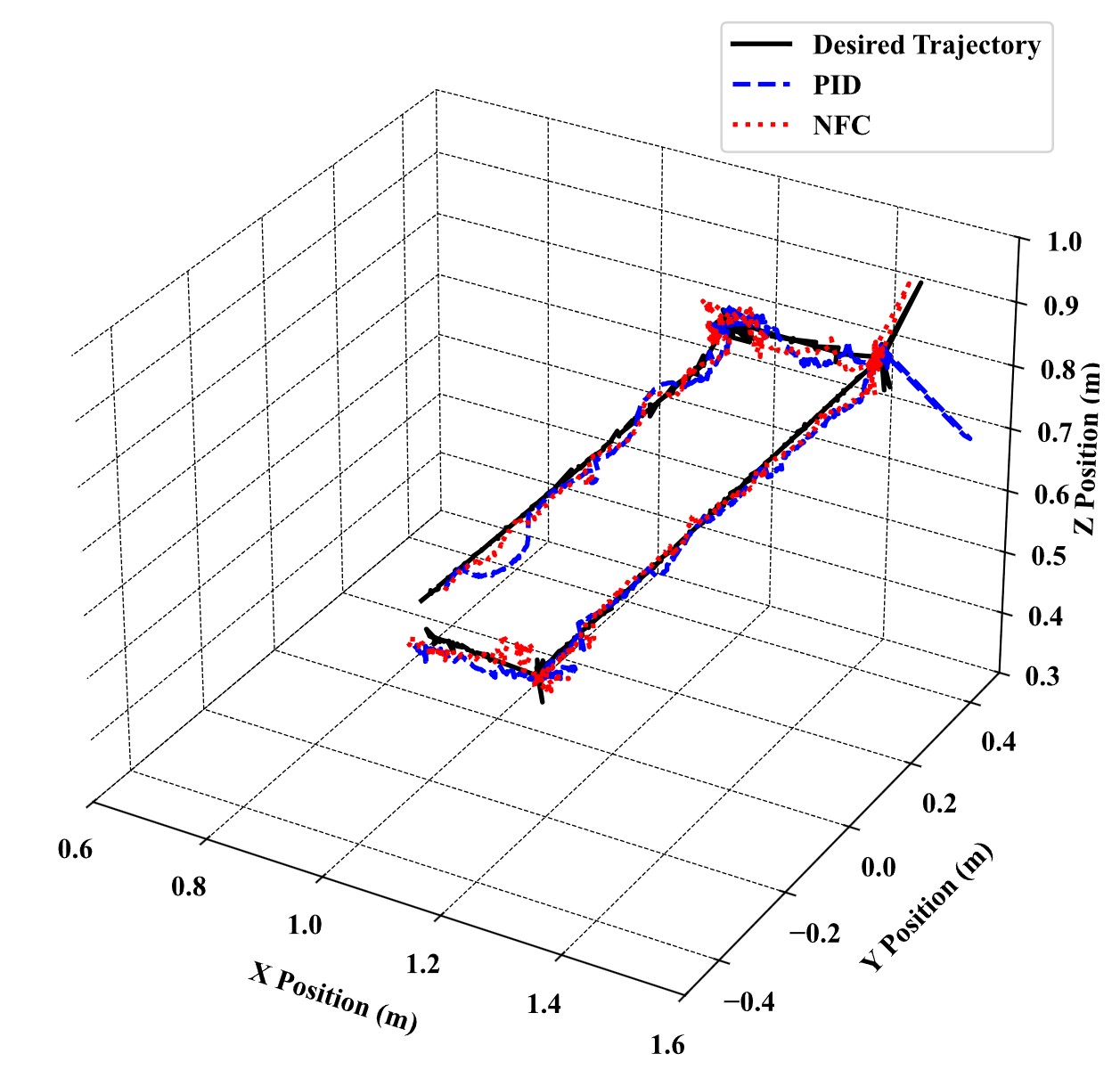}
        \caption*{(a)}
    \end{subfigure}
    \hfill
    \begin{subfigure}[b]{0.45\textwidth}
        \centering
        \includegraphics[width=0.7\textwidth]{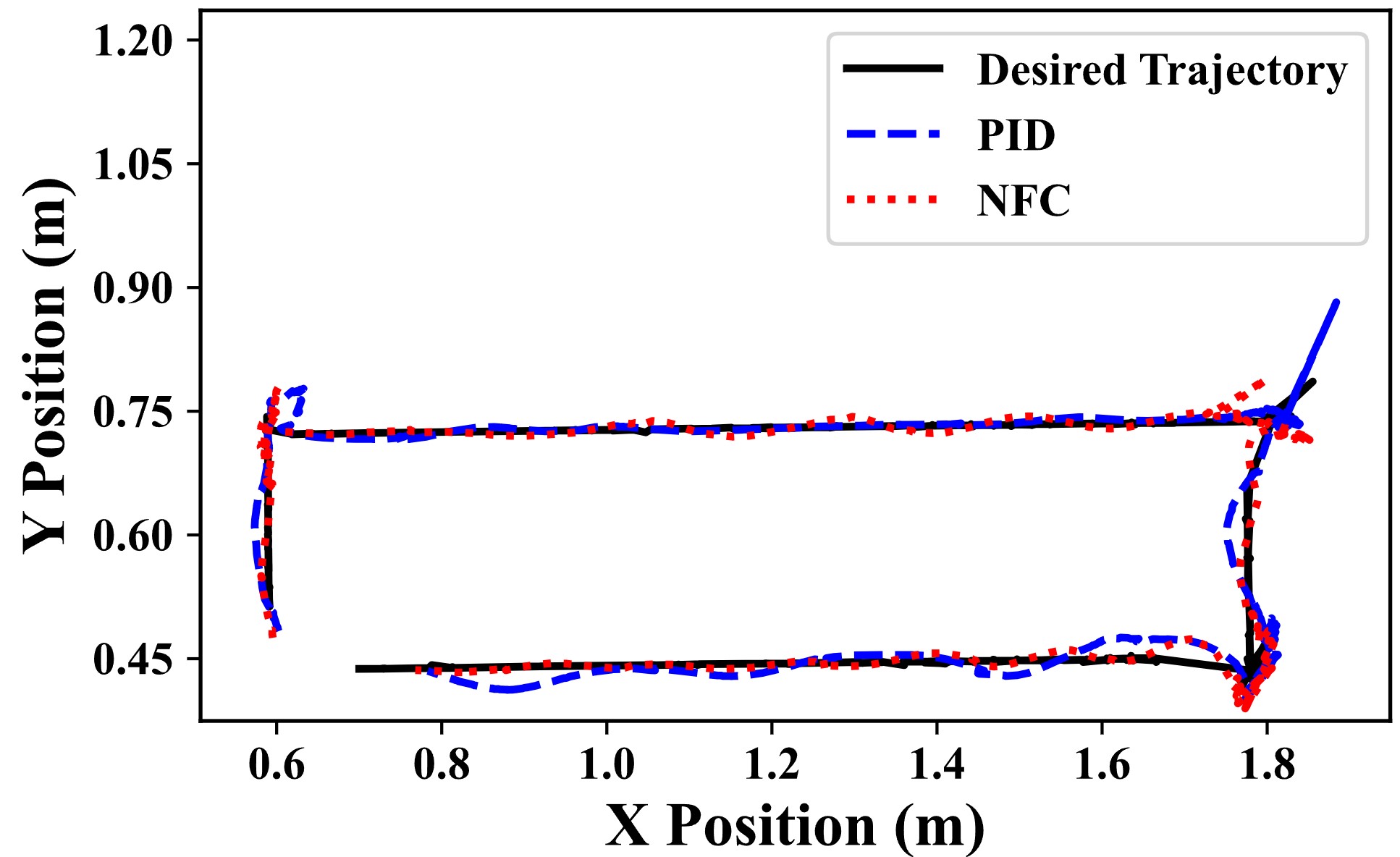}
        \caption*{(b)}
    \end{subfigure}
    \caption{Trajectory Tracking of PID control and NFC. (a) Trajectory in 3-D space. (b) Trajectory in the horizontal plane.}
    \label{fig:Trajectory}
\end{figure}

\begin{figure}[h!]

    \centering
    \begin{subfigure}[b]{\columnwidth}
        \centering
        \includegraphics[width=0.7\columnwidth]{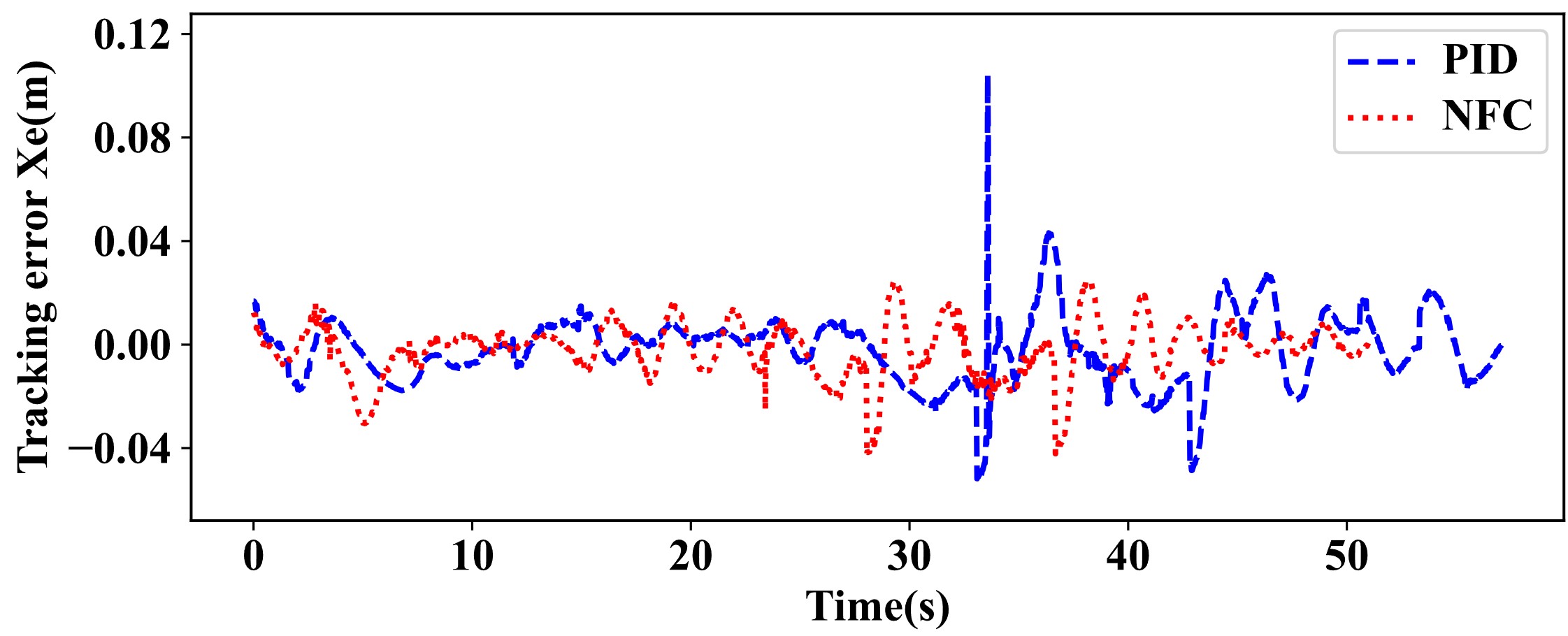}
        \caption*{(a)}
    \end{subfigure}
    \vspace{-1em} % 调整子图间的垂直间距
    \\
    \begin{subfigure}[b]{\columnwidth}
        \centering
        \includegraphics[width=0.7\columnwidth]{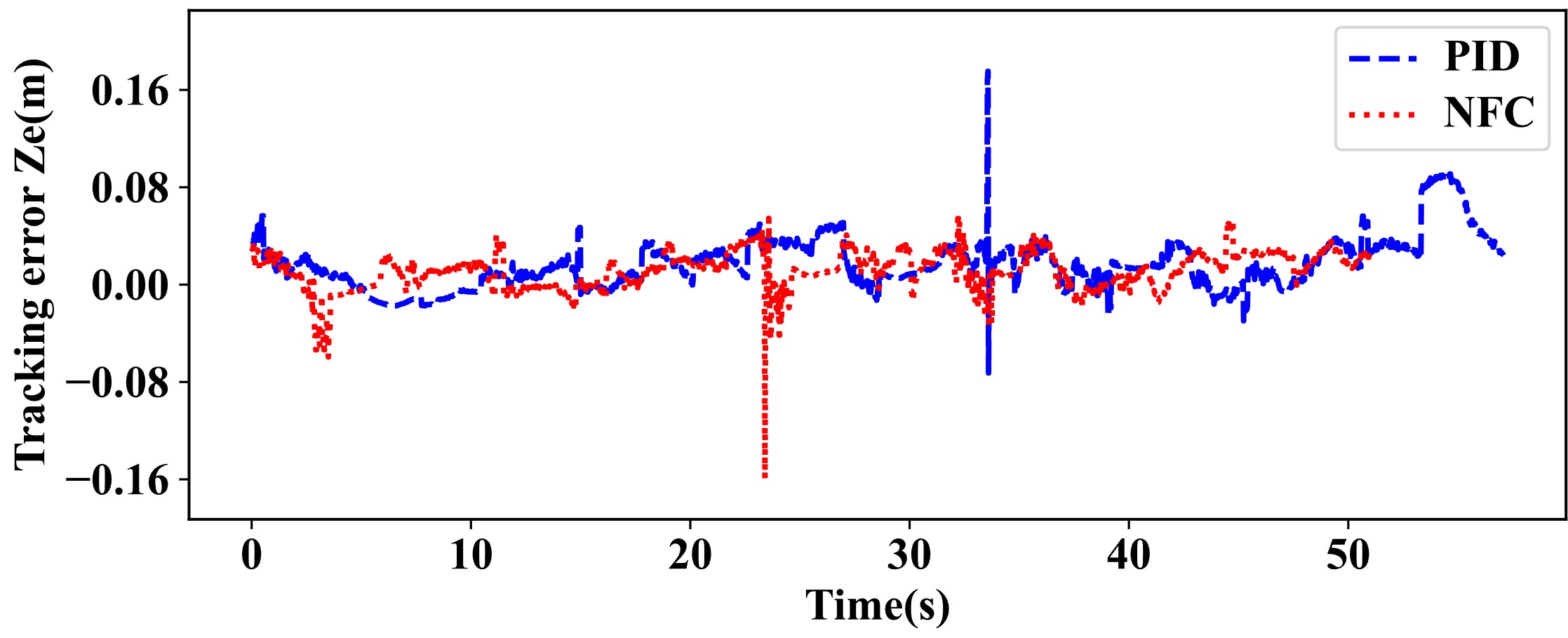}
        \caption*{(b)}
    \end{subfigure}
    \vspace{-1em} % 调整子图间的垂直间距
    \\
    \begin{subfigure}[b]{\columnwidth}
        \centering
        \includegraphics[width=0.7\columnwidth]{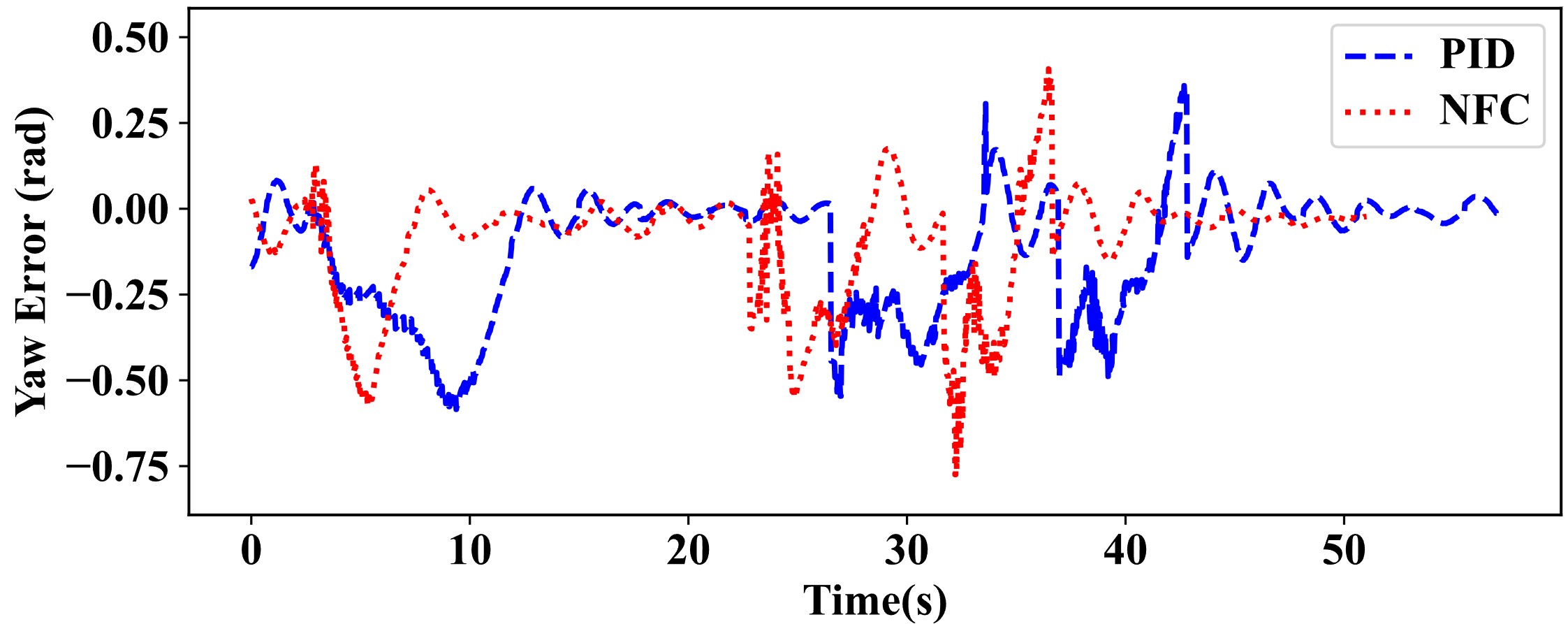}
        \caption*{(c)}
    \end{subfigure}
    \caption{Control errors of the proposed PID control and NFC. (a) Position control error \({x_e}\). (b) Position control error \({z_e}\). (c) Yaw control error \({\psi_e}\).}
    \label{fig:control_error}
\end{figure}

The experimental outcomes underscore critical aspects of the ROV's navigation system, including TagSLAM-based localization, path tracking, and overall control performance. In Figure~\ref{fig::tagslam_localization_result}, the enhanced localization achieved via TagSLAM and sensor fusion is clearly demonstrated. The left image shows the tags detected by the camera, while the right image depicts the robot's position on the map, emphasizing the effectiveness of sensor fusion for accurate navigation. Figure~\ref{fig:Trajectory} compares the actual trajectory of the ROV under PID control (dashed blue line), as well as NFC control (dotted red line) against the desired path (black line). Although both controllers closely follow the intended route, the NFC method provides superior precision and stability, particularly in three-dimensional movement and horizontal plane control. The Mean Average Error (MAE) of the three concerned dimension are 0.0081m (x-axis), 0.0157m (z-axis) and 0.120 rad (yaw angle) for NFC, while 0.011m, 0.0204m, 0.142rad for PID controller. The horizontal direction (y-axis) is not presented, as the ROV just move along y-axis in a fixed speed. Finally, Figure~\ref{fig:control_error} illustrates the tracking errors in position (X and Z axes) and yaw, where the NFC controller consistently exhibits lower and more stable errors compared to the PID controller, highlighting its enhanced capability to mitigate trajectory deviations and maintain stability.

\captionsetup[subfigure]{skip=-2.0pt} % 减小子图标题与图之间的间距

These results confirm that the proposed underwater inspection system enhances trajectory tracking and control accuracy with NFC, surpassing PID control, while the integration of TagSLAM ensures precise localization, making the system ideal for real-world underwater inspections.

\section{Conclusions}
This paper introduces an advanced ROV navigation system specifically designed for fishnet inspections, which enhances operational efficiency while reducing the reliance on manual labor. By employing ROS as the development platform, the system integrates a refined TagSLAM-based localization module, a dedicated path planning component, and an NFC-based control strategy. These innovations collectively contribute to marked improvements in the  localization accuracy and the trajectory tracking control. Experimental validations in a controlled laboratory setting confirm the system's ability to handle real-world challenges with precision and stability. Furthermore, its successful deployment on the navigation system underscores the practical applicability of this solution for future autonomous underwater inspections in dynamic environments.

\bibliography{ifacconf}             % bib file to produce the bibliography

\begin{thebibliography}{21}
\providecommand{\natexlab}[1]{#1}
\providecommand{\url}[1]{\texttt{#1}}
\providecommand{\urlprefix}{URL }
\expandafter\ifx\csname urlstyle\endcsname\relax
  \providecommand{\doi}[1]{doi:\discretionary{}{}{}#1}\else
  \providecommand{\doi}{doi:\discretionary{}{}{}\begingroup \urlstyle{rm}\Url}\fi

\bibitem[{Akram et~al.(2022)Akram, Casavola, Kapetanović, and Mišković}]{akram:22}
Akram, W., Casavola, A., Kapetanović, N., and Mišković, N. (2022).
\newblock A visual servoing scheme for autonomous aquaculture net pens inspection using rov.
\newblock \emph{Sensors}, 22(9), 3525.

\bibitem[{Amundsen et~al.(2021)Amundsen, Caharija, and Pettersen}]{amundsen:21}
Amundsen, H.B., Caharija, W., and Pettersen, K.Y. (2021).
\newblock Autonomous rov inspections of aquaculture net pens using dvl.
\newblock \emph{IEEE Journal of Oceanic Engineering}, 1--19.

\bibitem[{Amundsen et~al.(2024)Amundsen, Xanthidis, F{\o}re, Ohrem, and Kelasidi}]{AmuXanFor:24}
Amundsen, H.B., Xanthidis, M., F{\o}re, M., Ohrem, S.J., and Kelasidi, E. (2024).
\newblock Aquaculture field robotics: Applications, lessons learned and future prospects.
\newblock \emph{arXiv preprint arXiv:2404.12995}.

\bibitem[{Betancourt et~al.(2020)Betancourt, Coral, and Colorado}]{betancourt:20}
Betancourt, J., Coral, W., and Colorado, J. (2020).
\newblock An integrated rov solution for underwater net-cage inspection in fish farms using computer vision.
\newblock \emph{SN APPLIED SCIENCES}, 2(12), 1946.

\bibitem[{Cardaillac et~al.(2023)Cardaillac, Amundsen, Kelasidi, and Ludvigsen}]{cardaillac:23}
Cardaillac, A., Amundsen, H.B., Kelasidi, E., and Ludvigsen, M. (2023).
\newblock Application of maneuvering based control for autonomous inspection of aquaculture net pens.
\newblock In \emph{2023 8th Asia-Pacific Conference on Intelligent Robot Systems (ACIRS)}, 44--51.

\bibitem[{FAO(2024)}]{FAO:24}
FAO (2024).
\newblock The state of world fisheries and aquaculture 2024.
\newblock Accessed: October 31, 2024.

\bibitem[{Fossen(2011)}]{fossen:11}
Fossen, T.I. (2011).
\newblock Handbook of marine craft hydrodynamics and motion control.
\newblock \emph{John Wiley \& Sons Ltd}.

\bibitem[{Føre and Thorvaldsen(2021)}]{fore:21}
Føre, H.M. and Thorvaldsen, T. (2021).
\newblock Causal analysis of escape of atlantic salmon and rainbow trout from norwegian fish farms during 2010--2018.
\newblock \emph{Aquaculture}, 532, 736002.

\bibitem[{Haugaløkken et~al.(2023)Haugaløkken, Amundsen, Fadum, Gravdahl, and Ohrem}]{haugalokken2023adaptive}
Haugaløkken, B.O., Amundsen, H.B., Fadum, H.S., Gravdahl, J.T., and Ohrem, S.J. (2023).
\newblock Adaptive generalized super-twisting tracking control of an underwater vehicle.
\newblock In \emph{2023 IEEE Conference on Control Technology and Applications (CCTA)}, 687--693. IEEE.

\bibitem[{Leitmann(1981)}]{Leitmann:81}
Leitmann, G. (1981).
\newblock On the efficacy of nonlinear control in uncertain linear systems.
\newblock \emph{Journal of Nonlinear Control Systems}, 10, 1--5.

\bibitem[{López-Barajas et~al.(2023)López-Barajas, González, Sandoval, Gómez-Espinosa, Solis, Marín, and Sanz}]{lopez-barajas:23}
López-Barajas, S., González, J., Sandoval, P.J., Gómez-Espinosa, A., Solis, A., Marín, R., and Sanz, P.J. (2023).
\newblock Automatic visual inspection of a net for fish farms by means of robotic intelligence.
\newblock In \emph{OCEANS 2023 - Limerick}, 1--5.

\bibitem[{López-Barajas et~al.(2024)López-Barajas, Sanz, Marin-Prades, Gómez-Espinosa, González-Garcia, and Echague}]{lopez-barajas:24}
López-Barajas, S., Sanz, P.J., Marin-Prades, R., Gómez-Espinosa, A., González-Garcia, J., and Echague, J. (2024).
\newblock Inspection operations and hole detection in fish net cages through a hybrid underwater intervention system using deep learning techniques.
\newblock \emph{JOURNAL OF MARINE SCIENCE AND ENGINEERING}, 12(1), 80.

\bibitem[{Moore and Stouch(2014)}]{MooreStouchKeneralizedEkf2014}
Moore, T. and Stouch, D. (2014).
\newblock A generalized extended kalman filter implementation for the robot operating system.
\newblock In \emph{Proceedings of the 13th International Conference on Intelligent Autonomous Systems (IAS-13)}. Springer.

\bibitem[{Ohrem et~al.(2020)Ohrem, Kelasidi, and Bloecher}]{ohrem:20}
Ohrem, S.J., Kelasidi, E., and Bloecher, N. (2020).
\newblock Analysis of a novel autonomous underwater robot for biofouling prevention and inspection in fish farms.
\newblock In \emph{2020 28th Mediterranean Conference on Control and Automation (MED)}, 1002--1008.

\bibitem[{Pfrommer and Daniilidis(2019)}]{Pfrommer:19}
Pfrommer, B. and Daniilidis, K. (2019).
\newblock {TagSLAM}: Robust {SLAM} with fiducial markers.
\newblock \emph{arXiv preprint arXiv:1910.00679}.
\newblock ArXiv:1910.00679 [cs].

\bibitem[{Robotics(2025)}]{bluerov2}
Robotics, B. (2025).
\newblock Bluerov2 system components.
\newblock \url{https://bluerobotics.com/store/rov/bluerov2/}.
\newblock Accessed: 2025-02-12.

\bibitem[{Singh(1986)}]{Singh:86}
Singh, S.N. (1986).
\newblock Ultimate boundedness control of uncertain robotic systems.
\newblock \emph{International Journal of Systems Science}, 17(6), 859--863.

\bibitem[{Stenius et~al.(2022)Stenius, Folkesson, Bhat, Sprague, Ling, Ozkahraman, Bore, Cong, Severholt, Ljung, Arnwald, Torroba, Grondahl, and Thomas}]{stenius:22}
Stenius, I., Folkesson, J., Bhat, S., Sprague, C.I., Ling, L., Ozkahraman, O., Bore, N., Cong, Z., Severholt, J., Ljung, C., Arnwald, A., Torroba, I., Grondahl, F., and Thomas, J.B. (2022).
\newblock A system for autonomous seaweed farm inspection with an underwater robot.
\newblock \emph{SENSORS}, 22(13), 5064.

\bibitem[{Tani et~al.(2024)Tani, Ruscio, and Costanzi}]{tani:24}
Tani, S., Ruscio, F., and Costanzi, R. (2024).
\newblock Preliminary online validation of a visual-acoustic-based framework for autonomous underwater structure inspections.
\newblock In \emph{OCEANS 2024 - Halifax}, 1--8.

\bibitem[{Vasileiou et~al.(2024)Vasileiou, Manos, Vasilopoulos, Douma, and Kavallieratou}]{vasileiou:24}
Vasileiou, M., Manos, N., Vasilopoulos, N., Douma, A., and Kavallieratou, E. (2024).
\newblock Kalypso autonomous underwater vehicle: A 3d-printed underwater vehicle for inspection at fisheries.
\newblock \emph{Journal of Mechanisms and Robotics}, 16(4), 041003.

\bibitem[{Wang et~al.(2020)Wang, Ji, Liu, Tamura, Tsuchiya, Yamashita, and Asama}]{wang2020acmarker}
Wang, Y., Ji, Y., Liu, D., Tamura, Y., Tsuchiya, H., Yamashita, A., and Asama, H. (2020).
\newblock Acmarker: Acoustic camera-based fiducial marker system in underwater environment.
\newblock \emph{IEEE Robotics and Automation Letters}, 5(4), 5018--5025.

\end{thebibliography}
                                                     % with bibtex (preferred)

\end{document}